\documentclass{article}

\usepackage{microtype}
\usepackage{graphicx}
\usepackage{subfigure}
\usepackage{amsmath}
\usepackage{algpseudocode}
\usepackage{amssymb}
\usepackage{makecell}

\DeclareUnicodeCharacter{2032}{\ensuremath{\prime}}

\usepackage{booktabs} 

\usepackage{hyperref}

\usepackage[accepted]{icml2021}

\begin{document}

\twocolumn[
\icmltitle{SynthFormer: Equivariant Pharmacophore-based Generation of Synthesizable Molecules for Ligand-Based Drug Design}



\icmlsetsymbol{equal}{*}

\begin{icmlauthorlist}
\icmlauthor{Zygimantas Jocys}{to}
\icmlauthor{Zhanxing Zhu}{to}
\icmlauthor{Henriette M.G. Willems}{goo}
\icmlauthor{Katayoun Farrahi}{to}

\end{icmlauthorlist}

\icmlaffiliation{to}{Department of Vision Learning, and Control, Electronics and Computer Science, University of Southampton, Southampton, SO17 1BJ, United Kingdom}
\icmlaffiliation{goo}{The ALBORADA Drug Discovery Institute, University of Cambridge, Island Research Building, Cambridge Biomedical Campus, Hills Road, Cambridge, CB2 0AH, United Kingdom}

\icmlcorrespondingauthor{Zygimantas Jocys}{zj1g15@soton.ac.uk}

\icmlkeywords{Machine Learning, ICML}

\vskip 0.3in
]



\printAffiliationsAndNotice{}  

\begin{abstract}
Drug discovery is a complex, resource-intensive process requiring significant time and cost to bring new medicines to patients. Many generative models aim to accelerate drug discovery, but few produce synthetically accessible molecules. Conversely, synthesis-focused models do not leverage the 3D information crucial for effective drug design. We introduce SynthFormer, a novel machine learning model that generates fully synthesizable molecules, structured as synthetic trees, by introducing both 3D information and pharmacophores as input. SynthFormer features a 3D equivariant graph neural network to encode pharmacophores, followed by a Transformer-based synthesis-aware decoding mechanism for constructing synthetic trees as a sequence of tokens. It is a first-of-its-kind approach that could provide capabilities for designing active molecules based on pharmacophores, exploring the local synthesizable chemical space around hit molecules and optimizing their properties. We demonstrate its effectiveness through various challenging tasks, including designing active compounds for a range of proteins, performing hit expansion and optimizing molecular properties.
\end{abstract}
\vspace{-.6cm}

\section{Introduction}
Drug discovery is a complex, lengthy and expensive process \citep{drugcost}. Computer aided drug design has been shown to be effective \citep{schneider2005computer}, while generative machine learning (GML) methods offer a promising way to speed up the early stages of drug discovery by exploring molecules in an efficient and cost-effective way \citep{MEYERS20212707}. There are two main computational methods for early-stage drug discovery: de novo ligand design and virtual screening. State-of-the-art ML methods for target based molecule generation, such as TargetDiff \citep{guan20233dtargetdiff} and Pocket2mol \citep{Pocket2mol}, often overlook the practical aspects of synthesizing molecules in the lab. Conversely, screening models like Equibind \citep{stärk2022equibind} and Diffdock \citep{corso2023diffdock} are limited by the libraries they must screen. \textit{In vitro} screening is constrained by various factors, such as reactions, building blocks, and the methods used to obtain readings. DNA-encoded libraries (DELs) \citep{Brenner1992encodeddel}, high-throughput screening (HTS) \citep{inglese2007reportinghts}, and Dose-Response Curves \cite{crump1976fundamentaldrc} represent just a few of these approaches. HTS cannot be effectively optimized with machine learning, as its benefits arise from screening a fixed library, whereas DELs, while potentially optimized, are constrained by the building blocks and reactions used in their construction. Dose-response assays \citep{Crump1976Carcinogenicdoseresponse} are limited by low throughput. Therefore, it is crucial to integrate \textit{in vitro} technologies with \textit{in silico} techniques to optimize molecule discovery for each target and disease, thereby reducing costs and improving the speed of discovery.


One key failure point for generative ML methods in drug discovery is synthesisability.  Most methods rely on the Synthetic Accessibility (SA) score. However, it has been demonstrated that the SA score \citep{Ertl2009Estimation} is not a discriminative feature and is unable to determine whether a molecule is synthesizable, as shown in Synflownet \citep{cretu2024synflownetmoleculedesignguaranteed}. There have been multiple works that use a generative tool for a bottom up approach \citep{synnet} \citep{cretu2024synflownetmoleculedesignguaranteed}.  However, none of these models incorporate 3D information, which is essential for accurately understanding how molecules interact with their target protein, a critical factor in effective drug design.

Moreover, the most common molecule optimization approach relies on an external scorer blindly guiding the optimization as proposed by Guacamol \citep{Guacamol} and Moses~\citep{Moses}. However, in reality, the lead optimization  process is heavily constrained by the available synthesis paths which also become the basis for patenting the Markush structure \citep{markush}.

Pharmacophore-based ligand design has proven to be an effective approach for identifying early leads in drug discovery \cite{Dash2019}, particularly in cases where a protein structure with a defined binding mode is unavailable. When the structure of a protein target is not available or the mechanism of action is not well understood, protein folding algorithms, such as AlphaFold \cite{Jumper2021AlphaFold} or Boltz1 \cite{wohlwend2024boltz1}, or homology modelling may not produce a model suitable for structure-based design. In these cases, scientists often rely on ligand-based design, although this approach is limited by the need for multiple ligands with affinity to the target protein as input. This underscores the need for models capable of generating diverse compound libraries that can be screened based on general hypotheses.

To address these challenges we propose the SynthFormer, a model that uses 3D equivarient encoder for pharmacophores and generates fully synthesisable molecules by building them as synthetic trees. We validate the effectiveness of the model by showcasing its ability to generate molecules that have good docking scores for a variety of proteins, as well as show that this tool can be effective for molecule optimisation. The main contributions of our work
can be summarized as follows:
\begin{itemize}
    \item We introduce the first methodology that incorporates both 3D information and synthesis to make pharmacophore ligand design possible.
    \item We propose a novel Synthesis-Aware  Decoder, called \textit{SynthFormer}, for generating  molecules by translating 3D pharmacophore representations into synthesizable compounds.
    \item We demonstrate that the SynthFormer model is capable of generating molecules with strong docking performance, performing hit expansion, and can be used for molecule optimization.
\end{itemize}

\section{Related work}

\paragraph{Combinatorial Optimization}{
Approaches in combinatorial optimization employ a range of chemical building blocks and reaction templates to construct a combinatorial chemical space, such as Galileo \citep{meyenburg2023galileo3dlarge}. These strategies then apply optimization techniques, such as genetic algorithms \citep{Gao2021} and Monte Carlo tree search \citep{Swanson2023}, to explore this space for desired molecules. Prior to the advent of molecular deep learning, early efforts like SYNOPSIS \citep{Vinkers2003} generated candidate molecules through virtual reactions, selecting them based on scoring functions. Recent advancements in deep learning for predicting chemical reactions \citep{Coley2019a} have further inspired methods that utilize neural networks to forecast reaction outcomes without relying on predefined templates. }

\paragraph{Synthesis-Aware Deep Learning} {Related work spans various approaches to molecule design, based on reinforcement learning (RL) or self-supervised learning. RL-based methods, such as SynFlowNet \cite{cretu2024synflownetmoleculedesignguaranteed} and SynNet \cite{synnet}, optimize molecular properties by simulating sequential decisions, constrained by synthesizability and reaction feasibility. The core limitation is that the reward function has limited accuracy. The Synflownet model was guided by the docking module, which has an accuracy of 0.41. These methods rely on reward functions tailored to specific objectives, such as activity or docking scores. On the other hand, self-supervised learning approaches, like ChemProjector \cite{luo2024projectingmoleculessynthesizablechemical}, learn meaningful molecular representations by predicting structural or functional transformations in a molecule-to-molecule framework. However, this does not allow to model activity cliffs \cite{zhang2023activitycliffpredictiondataset}, as it is only transforming the molecules based on the graph and has no binding information.}

\paragraph{3D Generative Models} {There are two primary approaches for shape-based generation: conditioning on a protein structure or on a known active ligand. Structure-based methods, such as TargetDiff \citep{guan20233dtargetdiff} and Pocket2Mol \citep{Pocket2mol}, are designed to generate molecules that fit within a protein binding site. However, benchmark studies \citep{Drugpose} have demonstrated that while these methods can produce molecules that theoretically fit, they often fail to produce synthesis-ready compounds\citep{luo2024projectingmoleculessynthesizablechemical} and may not bind as intended. Ligand-based approaches, like SQUID \citep{SQUID} and LigDream \citep{ligdream}, encode the shape or pharmacophores, and use another decoder to decode the molecules. Despite their promise, these networks have also struggled with generating synthesisable molecules.}

\paragraph{Molecule Optimization} {Molecule optimization in a machine learning context, as described in Guacamol \citep{Guacamol} and Moses \citep{Moses}, involves improving molecules by using scoring functions to evaluate how well they meet desired property profiles. The objective is to generate molecules that maximize these scores, while decoupling the optimization process from the challenge of selecting appropriate scoring functions \citep{schneider2005computer}. This approach has been adopted by the ML community for various representations, including SMILES-based \citep{decao2022molgan}, graph-based without 3D information, reinforcement learning methods \citep{cretu2024synflownetmoleculedesignguaranteed}, and 3D approaches \citep{Pocket2mol}. At the same time, this optimization framing is detached from the drug discovery process in real life. In drug discovery, once Hits are identified and Structure-Activity Relationship (SAR) is established, lead optimization involves exploiting the known synthetic pathways and exploring similar routes to improve the molecule. This was done in this lead optimization study \citep{leadop}, where they find and optimize inhibitors of RIP1 Kinase. }

\paragraph{Limitations of Existing Methods} {Combinatorial optimization techniques are constrained by their reliance on predefined reaction templates, which result in a slow exploration of chemical space. On the other hand, deep learning models typically focus on only one part of the task: generating structurally and physically accurate structures or synthetically accessible compounds, but not both. Moreover, while deep learning methods excel in these areas, they often fail to integrate the optimization process with the specific needs of lead optimization. As a result, the generated compounds may not align well with the practical requirements of drug discovery, such as feasibility of synthesis or sufficient solubility for assay conditions. This detachment between structural design and practical application can hinder the overall success of deep learning approaches in optimizing drug candidates.}

{SynthFormer is a model designed to address these challenges. To the best of our knowledge this is the first model to translate 3D geometric profiles of pharmacophores and constructs molecules from available building blocks based on specified requirements, only operating in the plausible chemical space. Additionally, its architecture operates at the synthetic level, enabling an optimization process that mimics the work of a chemist in real drug discovery scenarios.}

\section{Method}
\begin{figure*}[h!]
    \centering
    \includegraphics[width=\textwidth, page=1]{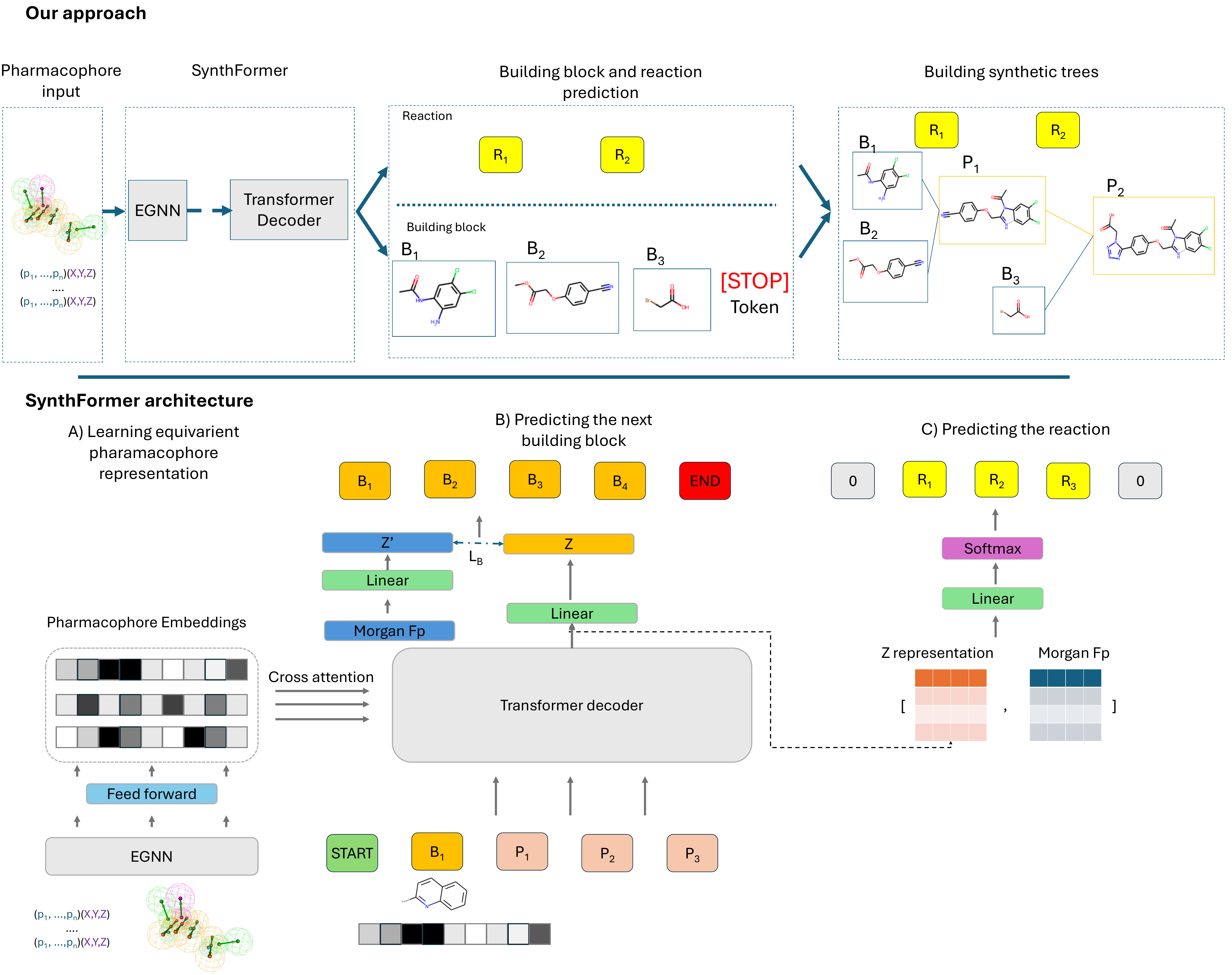}
    \vspace{-.8cm}
    \caption{The generation process begins by representing the pharmacophores as a fully connected graph, passed through an EGNN to obtain pharmacophore embeddings. These embeddings are then propagated to a transformer decoder. The transformer decoder first receives a start token and predicts building block \( B_1 \), followed by reaction \( R_0 \). Next, it takes the fingerprint of \( B_1 \) as input to the decoder and predicts building block \( B_2 \), followed by reaction \( R_1 \), which is applied to generate product \( P_1 \). \( P_1 \) is then used to predict building block \( B_3 \) and reaction \( R_2 \), producing product \( P_3 \). This process repeats until the end token is generated.}
    \vspace{-.4cm}
    \label{fig:Synthprocess}
\end{figure*}

SynthFormer introduces a novel methodology for molecule generation, characterized by innovative data preparation, novel input representations, and a unique architectural synergy between Equivariant Graph Neural Networks (EGNNs) and Synthesis-Aware  Decoder (see Sec.~\ref{sec:arch} for both), as sketched in Figure~\ref{fig:Synthprocess}. Our approach leverages a new type of input: desired pharmacophore features paired with their 3D coordinates. These inputs are encoded by the EGNN into rich embeddings that comprehensively capture critical spatial properties, providing a deeper understanding of molecular structure. These embeddings are subsequently processed by a transformer decoder, which autoregressively predicts a sequence of building blocks and reactions, ensuring a smooth and interpretable generation process. 

To train and evaluate our model, we need to generate a custom dataset. Existing deep learning models for chemistry often operate at the atom level and rely on datasets like PDBBind \cite{wang2005pdbbind}, QM9 \cite{wu2018moleculenetbenchmarkmolecularmachineqm9}, or QMugs \cite{Isert2022QMugs}. However, since our approach focuses on pharmacophores and their synthetic trees—and because we rely on Enamine's extensive chemical database (to which they graciously provided access)—it is necessary to construct our own dataset. This dataset will support the development and testing of the first model specifically designed to operate on pharmacophores in the context of synthetic tree generation, as elaborated in the following part.

\subsection{Encoder Input: Pharmacophore}
For data preparation, we construct input batches by combining pharmacophore features and 3D spatial information derived from conformers. Pharmacophores, represented as one-hot encoded vectors, are assigned to each atom, while the conformers provide the corresponding \((x, y, z)\) coordinates. Together, these are used to create the input data for the EGNN encoder, ensuring it captures both the spatial geometry and the pharmacophore profile of the molecule.

\textbf{Conformers}
A molecule conformer is a spatial arrangement of atoms in a molecule that can be rotated around single bonds. Consider a molecule as a graph \(G = (\mathcal{V}, \mathcal{E})\) with atoms \(v \in \mathcal{V}\) and bonds \(e \in \mathcal{E}\), and denote the space of its possible conformers \(\mathcal{C}_G\). A conformer \(C \in \mathcal{C}_G\) can be specified in terms of its \textit{intrinsic} (or internal) coordinates: local structures \(L\) consisting of bond lengths, bond angles, and cycle conformations; and torsion angles \(\tau\) consisting of dihedral angles around freely rotatable bonds (precise definitions in Appendix A). For each molecule, the coordinates can be represented as a set \(\{ \mathbf{x}_v^k \in \mathbb{R}^3 \mid v \in \mathcal{V}, k \in \{1, 2, \ldots, n_v\} \}\), where \(\mathbf{x}_v^k = (x_v^k, y_v^k, z_v^k)\) represents the \(k\)-th set of 3D coordinates for atom \(v\), and \(n_v\) is the number of possible solutions for the coordinates of atom \(v\).

\paragraph{Pharmacophores} are abstract representations of the molecular features necessary for the molecular recognition of a ligand by a biological macromolecule, which are crucial for its biological activity. We selected the following pharmacophores types for one-hot encoding available in the RDKit \citep{RDKit} package: Hydrogen Bond Donors (HBD), Hydrogen Bond Acceptors (HBA), Aromatic Rings (AR), Hydrophobic Centers (HC), Positive Ionizable features (PIF), and Negative Ionizable features (NIF). In our approach, we represent pharmacophores through a one-hot encoding scheme. Specifically, we denote a pharmacophore feature vector as \( \mathbf{p} = [p_1, p_2, \ldots, p_n] \), where each \( p_i \in \{0, 1\} \) of the pharmacophore type.

\paragraph{Data Preparation}
To generate the data, we begin by selecting a set of building blocks and perform a sequence of  $n$ random reactions from the available set, saving the resulting products. Next, we repeat the process but include the option to sample both the initial molecules and the previously generated products, effectively exploring the chemical space through random sampling. Once the molecules are sampled, we compute a random conformer for each one using Merck Molecular Force Field (MMFF) optimization from RDKit \citep{RDKit}. Afterwards, we extract the pharmacophores and 3D coordinates of each molecule. This results in an input dataset where the pharmacophores and 3D coordinates serve as the inputs, and the output prediction corresponds to the synthetic tree.

\subsection{Decoder Output: Molecules as Synthetic Trees}
A molecule is generated by starting from the building blocks and iteratively applying the reactions. There are some cases where a reaction has multiple possible main products. For ease of discussion, we assume that reactions produce only one main product in this section. 

As illustrated by the right panel of the first row in Figure~\ref{fig:Synthprocess},  each node in the tree represents a compound, while the edges denote the reactions that transform one compound into another. This structure is useful for predicting feasible synthetic routes, optimizing reaction pathways, and planning the step-by-step construction of complex molecules.

For example, we begin with building block \( B_1 \), then we obtain the next building block, \( B_2 \) with the first reaction, \( R_1 \), that transforms \( B_1 \) into \( P_1 \). This reaction generates product \( P_1 \), thereafter we acquire the next building block, \( B_3 \), and the second reaction, \( R_2 \), which transforms \( P_1 \) into \( P_3 \). This process continues, step by step until the target molecule is reached, with each product serving as input for the next reaction, forming a complete synthetic tree.

Moreover, when sampling a batch, we can select any molecule as the final molecule and trace all the previous steps leading to the desired molecule, allowing for efficient exploration.

\subsection{Architecture}
\label{sec:arch}
SynthFormer is a two-part model that combines pharmacophore features with 3D spatial information from conformers to generate new molecules from building blocks. By integrating one-hot encoded pharmacophore vectors with conformer-derived coordinates, the EGNN encoder processes this data to generate rich embeddings. These embeddings are then passed to a transformer decoder that autoregressively generates building blocks and reactions, ensuring the synthesis of molecules with the desired pharmacophore features and synthetic accessibility.

\textbf{Equivariant Pharmacophore Encoder}
We employ Cartesian coordinate features to preserve equivariance with the pharmacophore representation p to encode the pharmacophoric profile. We treat this as a fully connected graph, because our hypothesis space is not limited by any bonds. We use the EGNN \citep{satorras2021en}, that takes as input the set of node embeddings $\mathbf{p} = \{\mathbf{p}_0, \ldots, \mathbf{p}_{M-1}\}$, coordinate embeddings $\mathbf{x}' = \{\mathbf{x}_0', \ldots, \mathbf{x}_{M-1}'\}$, and edge information $\mathcal{E} = (e_{ij})$ and outputs a transformation on $\mathbf{h}^{l+1}$ and $\mathbf{x}'^{l+1}$. The equations that define this layer are the following:
\begin{equation}
    m_{ij} = \phi_e\left( \mathbf{h}_i^l, \mathbf{h}_j^l, \left\|\mathbf{x}_i' - \mathbf{x}_j'\right\|_2^2, e_{ij} \right) 
    \tag{1} \label{eq:message_function}
\end{equation}
\begin{equation}
    \mathbf{x}_i'^{l+1} = \mathbf{x}_i' + \sum_{j \neq i} (\mathbf{x}_i' - \mathbf{x}_j') \phi_x (m_{ij}) 
    \tag{2} \label{eq:position_update}
\end{equation}
\begin{equation}
    m_i = \sum_{j \in \mathcal{N}(i)} m_{ij} 
    \tag{3} \label{eq:message_aggregation}
\end{equation}
\begin{equation}
    \mathbf{h}_i^{l+1} = \phi_h \left( \mathbf{h}_i^l, m_i \right) 
    \tag{4} \label{eq:node_update}
\end{equation}
 In Eq. \ref{eq:message_function} , we input the relative squared distance between two coordinates $\left\|\mathbf{x}_i' -\mathbf{x}_j'\right\|_2^2$ into the edge operation $\phi_e$. Later, we use the representation \(\mathbf{h}^{l+1}\) and pass it through cross-attention to the decoder.Finally, \( \varphi_e \) and \( \varphi_h \) are the edge and node operations, respectively, which are commonly approximated by Multilayer Perceptrons (MLPs).
 We employ a seven-layer EGNN as the encoder in our model.

\textbf{Synthesis-Aware Decoder}
The input to the transformer begins with a token $[START]$, followed by the first B, then $P_1$. The input represents the last final product from the reactions. The model stops when the building block MLP predicts the end token $[END]$. An input token is a Morgan fingerprint and start token, denoted by $[f_p, start]$, where $f_p \in \{0, 1\}^{4096}$ is the Morgan fingerprint of length 4096 and radius 3 \citep{Morgan1965}. Morgan fingerprint encodes local atomic environments by applying a circular algorithm that captures patterns in the arrangement and connectivity of atoms within a specified radius around each atom. This algorithm considers features such as atomic types, bonding patterns, and the functional groups in the neighborhood of each atom. The result is a fixed-length bit vector that summarizes these structural patterns. Building block tokens are converted into embeddings by an MLP. Moreover, we use positional encodings (PE) to encode the position for each molecule.
The decoder uses the standard multihead attention mechanism. However, at the output of the decoder, we take the multihead output pass and concatenate with the learnable molecular embeddings.
\begin{equation}
\text{MultiHead}(Q, K, V)=\text{Concat}(\text{h}_1,..., \text{h}_h)W^O
\end{equation}
\begin{equation}
\text{where h}_i = \text{Attention}(QW_i^Q, KW_i^K, VW_i^V),
\end{equation}
and the projections are parameter matrices \( W_i^Q \in \mathbb{R}^{d_{\text{model}} \times d_k}, W_i^K \in \mathbb{R}^{d_{\text{model}} \times d_k}, W_i^V \in \mathbb{R}^{d_{\text{model}} \times d_v}, \) and \( W^O \in \mathbb{R}^{h d_v \times d_{\text{model}}} \).
The encoder is composed of a stack of \( N = 7 \) identical layers and we employ \( h = 8 \) attention heads. We pass the input tokens through the last multi-head attention mechanism to obtain the representation Z. During training, we utilize cosine similarity to guide the model towards generating molecules with desired properties. This involves calculating the cosine similarity between the generated molecule's representation ($Z'$) and the target molecule's representation ($Z$), and using this similarity score as a component of the loss function. During sampling, we leverage the trained model to generate new molecules by maximizing the cosine similarity between the generated molecule's representation and the representation embeddings, effectively guiding the generation process towards molecules that align with the learned chemical space:
\begin{equation}
B_{i+1} = argmax( \text{Similarity}\left(Z_i, Z') \right))
\end{equation}
Next, we concatenate the representation \( Z_i \) with the embeddings of the predicted building block \( B_{i+1} \) and apply a linear transformation followed by a softmax function to predict the reaction \( R_{i+1} \). This process can be expressed as:
\begin{equation}
R_{i+1} \sim  \text{Softmax}(\text{Linear}(\text{Concat}(Z_i, \text{f}_{p}(B_{i+1}))))
\end{equation}
The model uses both the predicted building block \( B_{i+1} \) and the reaction \( R_{i+1} \) to generate the product, which is then fed back into the model as input for the next iteration. This process continues iteratively, generating new building blocks and reactions, until the model predicts an end token for the building block, signaling the completion of the process.

\subsection{Training}
In our study, we employ causal masking within the transformer architecture to ensure that the model adheres to the autoregressive property during training and inference. Causal masking is crucial for preventing the model from accessing future tokens in the sequence when predicting the current token, thereby maintaining the integrity of the temporal or sequential data flow.

The building block loss function for the building block fingerprint, where \(\ell\) is the length of the sequence, is defined:
\begin{equation}
L_{\text{B}} =  \frac{1}{n} \sum_{j=0}^{l-1} \frac{Z'_{i+1} \cdot Z_{i+1}}{||Z'_{i+1}|| \cdot ||Z_{i+1}||} \label{eq:Lb}
\end{equation}
where \(Z\) and \(Z'\) are vectors encoding the \(Z\) representation and the encoded building block representation, respectively. The encoded Morgan fingerprint representation \(Z'\) is given:
\begin{equation}
Z' = \mathbf{W} \cdot \text{f}_{p}(B_{i+1}) + \mathbf{b} \label{eq:Zprime}
\end{equation}
The reaction loss function \(L_{\text{rxn}}\) is defined:
\begin{equation}
L_{\text{rxn}} = \frac{1}{m} \sum_{i=0}^{\ell-1} \text{CE} (\hat{r}_{i+1}, r_{i+1}) \label{eq:Lrxn}
\end{equation}
The final loss is the sum of the two terms defined:
\begin{equation}
L = L_{\text{B}} + L_{\text{rxn}} \label{eq:final_loss}
\end{equation}
\vspace{-1cm}
\subsection{Inference}
At inference time, we prepare the pharmacophore data and the start token as inputs. The transformer decoder first receives a start token and predicts building block \( B_1 \), followed by reaction \( R_0 \). Next, it takes the fingerprint of \( B_1 \) as input and predicts \( B_2 \), followed by reaction \( R_1 \), which is applied to generate product \( P_1 \). Product \( P_1 \) is then used to predict building block \( B_3 \) and reaction \( R_2 \), producing product \( P_3 \). This process repeats until the end token is generated. Finally, an end token $[END]$ terminates the
process and the last molecule is returned as the product.


\section{Experiments}
In this section, we mainly examine the performance of our model in the context of active ligand design. First, we compare the model against a random baseline to assess its ability to generate biologically relevant molecules. Next, we evaluate its performance against synthesis-aware models, highlighting its advantages in generating molecules that are diverse and active. We further compare our approach with three ligand-based 3D models, focusing on its docking performance and its unique ability to generate 100\% synthesisable molecules. Moving beyond ligand design, we explore the utility of embedding encodings, by generating new analogs with different properties. Finally, we demonstrate the model’s potential for molecular optimization.
\subsection{Experimental Setup}
\textbf{Reaction Templates } 
We use the publicly available reaction template set \citep{Hartenfeller}. The template set contains 58 reaction templates encoded in SMARTS format.

\textbf{Building Blocks }{
We use commercially available building blocks (BBs) from Enamine \cite{enamine}, which are molecules prepared in bulk to be readily synthesised into candidate molecules. There are 251,222 building blocks.}

\textbf{Similarity Scores }
To evaluate the similarity between the input molecule and the output molecule, we use the Tanimoto similarity score on three different fingerprints: (1) Morgan fingerprint of length 4096 and radius 2 \citep{Morgan1965}, (2) Morgan fingerprint of Murcko scaffold \citep{murko}. The three similarity scores indicate chemical similarities in three different aspects: overall structure, scaffold structure, and pharmacophore property, respectively, and are all normalized to [0, 1].

\textbf{Docking }{We used SMINA \cite{smina} for docking, converting ligands and proteins to PDBQT format with OpenBabel \cite{openbabel}. The docking box, centered on each ligand's centroid, was set to 25x25x25 Å. Up to 10 poses were generated for each ligand docked into the protein’s binding site. This process was repeated for all ligand-protein pairs for further analysis.}
\vspace{-.2cm}
\subsection{Main Evaluation: Designing Active Compounds}

\begin{figure*}
    \centering
    \includegraphics[width=\textwidth]{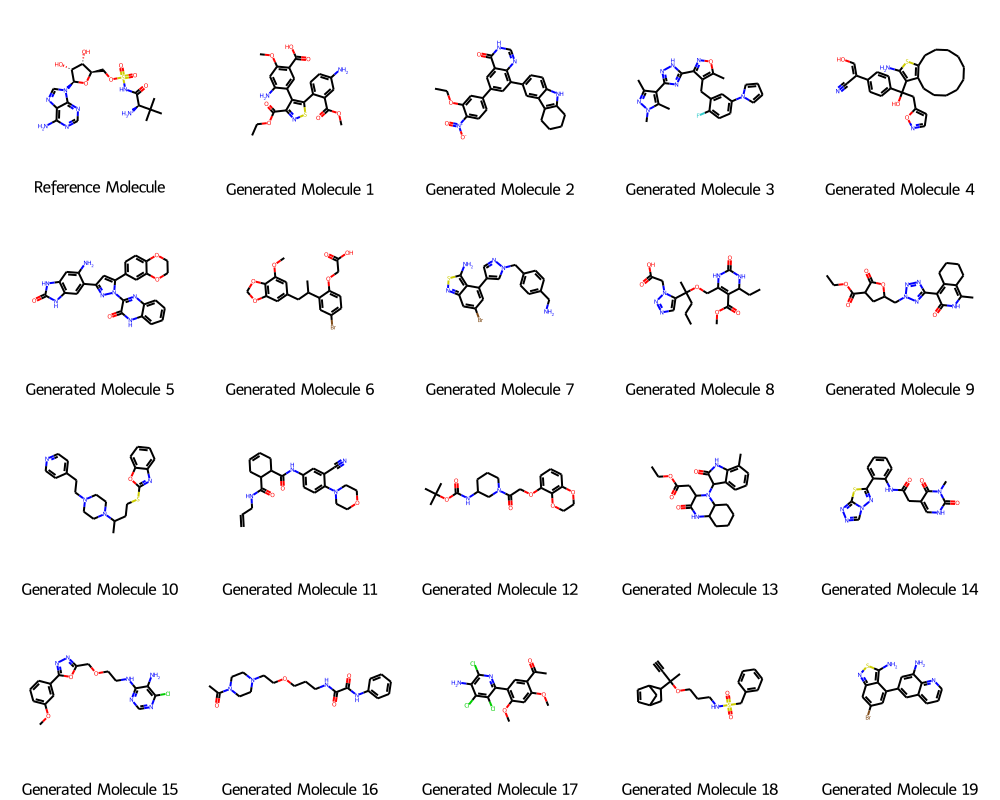}  
    \caption{The first molecule corresponds to the reference structure derived from the 3COY PDB ID. The subsequent molecules are computationally generated using the SynthFormer model, illustrating its capability to design novel molecular structures based on a known protein-ligand complex. These results highlight SynthFormer's potential in generating diverse and plausible molecular candidates.}
    \label{fig:SynthFormer_active_molecules}
\end{figure*}

Structural similarities between the generated and original compounds were assessed, followed by redocking of the generated compounds. Docking scores were compared to those of the original ligands to evaluate binding affinity and pose. Notably, for PDB entries 1x8d, 2bt9, 416d, and 5fl4, the average docking energy closely matched the reference, with the lowest docking energy consistently outperforming the reference, except for 3ga5 (see Table \ref{tab:compare against random}). Mean docking energies for generated molecules ranged from -4.55 to -10.14 kcal/mol (Appendix A to see similar molecules, Figure \ref{fig:SynthFormer_active_molecules}).
\vspace{-.4cm}
\begin{table}[H]
    \centering
    \begin{tabular}{@{}lcccccc@{}}
        \toprule
        PDB ID &  MW (g/mol) & LogP 5\% & LogP 95\% &  QED \\
        \midrule
        1x8d & 320.25 & 0.27 & 5.14 & 0.59 \\
        1xbo & 313.36 & 0.00 & 4.09 & 0.60 \\
        2afw & 320.32 & -0.01 & 5.41 & 0.58 \\
        2aog & 320.53 & 0.63 & 4.73 & 0.60 \\
        2bt9 & 305.96 & 0.02 & 4.60 & 0.60 \\
        3coy & 313.24 & 0.12 & 4.72 & 0.59 \\
        3ga5 & 318.71 & 0.24 & 5.13 & 0.60 \\
        4q6d & 331.75 & -0.10 & 5.23 & 0.57 \\
        5fl4 & 319.19 & 0.14 & 4.77 & 0.58 \\
        5ka1 & 339.02 & -0.41 & 4.80 & 0.56 \\
        \bottomrule
    \end{tabular}
    \vspace{-0.25cm}
    \caption{The generated molecules exhibit diverse molecular weights (305.96–339.02 g/mol), LogP values within the Ghose filter range (-0.4 to 5.6), and reasonable drug-likeness (QED).}
  \label{tab:mol_properties}    
  \vspace{-.3cm}
\end{table}

\begin{table*}[t]
\centering
\begin{tabular}{|c||c|c|c||c|c|c|}
\hline
\textbf{PDBID} & \textbf{Ref Dock} (kcal/mol) & \makecell{\textbf{Av. Dock Gen} \\ \textbf{(Min)} (kcal/mol)$\downarrow$} & \makecell{\textbf{Random} \\ \textbf{Baseline} (kcal/mol)} & \textbf{Murcko}$\downarrow$ & \textbf{Tanimoto}$\downarrow$ \\
\hline
1x8d & -6.17 & -6.43 (-8.33) & -4.62 & 0.07 & 0.10 \\
1xbo & -10.79 & -8.22 (-11.22) & -3.75 & 0.11 & 0.12 \\
2afw & -3.85 & -7.92 (-9.21) & -4.96 & 0.08 & 0.09 \\
2aog & -10.45 & -8.63 (-11.53) & -4.13 & 0.01 & 0.07 \\
2bt9 & -6.68 & -7.63 (-10.11) & -5.72 & 0.03 & 0.08 \\
3coy & -12.41 & -10.14 (-12.26) & -4.08 & 0.07 & 0.08 \\
3ga5 & -9.68 & -4.55 (-7.60) & -4.62 & 0.04 & 0.05 \\
4q6d & -7.41 & -7.55 (-10.07) & -3.52 & 0.10 & 0.07 \\
5fl4 & -8.09 & -8.01 (-9.44) & -3.94 & 0.12 & 0.11 \\
5ka1 & -7.67 & -6.44 (-9.71) & -5.21 & 0.07 & 0.07 \\
\hline
\end{tabular}
\vspace{-0.25cm}
\caption{The table presents docking energies of the reference compound from the PDBID (Ref Dock) and the mean and minimum docking energies (Av. Dock Gen (Min)) for 100 generated molecules docked to the same protein as well as the Random Baseline from Enamine space. Tanimoto and Murcko similarities reflect the average resemblance between the 100 generated molecules and the reference ligands.}
\label{tab:compare against random}
\vspace{-0.4cm}
\end{table*}
The similarity metrics, Tanimoto and Murcko, provide insights into the structural resemblance between the generated molecules and reference ligands indicating extremely low similarity, ranging from 0.06 to 0.12. As shown in Table \ref{tab:mol_properties}, the generated molecules adhere to the Ghose filter, with molecular weights ranging from 305.96 to 339.02 g/mol, LogP values within the acceptable range (-0.4 to 5.6), and reasonable drug-likeness.

\begin{table*}
\centering
\label{tab:comparison_combined}
\begin{tabular}{|c|c|c|c|c|}
\hline
\textbf{Method}      & \textbf{Type}       & \textbf{$\Delta$ Dock} (kcal/mol) $\downarrow$ & \textbf{Tanimoto} $\downarrow$ & \textbf{Synthesis}$\uparrow$ \\ \hline
ChemProjector           & Synthesis           & 2.55                               & 0.53                          & 100\%                          \\ \hline
SynNet               & Synthesis           & 2.7                                & 0.64                          & 100\%                          \\ \hline \hline
SQUID                & 3D                  & \textbf{2.39}                      & 0.24                          & 23.6\%                         \\ \hline
Ligdream             & 3D                  & 2.78                               & 0.22                          & 32.4\%                         \\ \hline \hline
SynthFormer          & Synthesis+3D        & 2.46                               & \textbf{0.09}                 & 100\%                          \\ \hline
\end{tabular}
\vspace{-0.25cm}
\caption{Comparison of synthesis-aware, 3D generative methods and SynthFormer for generating molecules across 10 PDB (100 compounds per PDB). SynthFormer achieves the lowest similarity to reference ligands while maintaining docking performance comparable to 3D generative models like SQUID. While, it generates more diverse and more active compounds than ChemProjector and SynNet.}
\label{tab:comparemodels}
\vspace{-0.4cm}
\end{table*}

\paragraph{Comparison to Random Baseline}{
In our experiments, as shown in Table \ref{tab:compare against random}, the best generated molecule  with the best bose consistently outperforms the reference molecule in terms of docking energy. This result is particularly notable as we observe lower docking energies for the generated molecules across all cases compared to the reference, highlighting the superior performance of our approach. Furthermore, when compared to the random baseline, our generated molecules consistently beat the baseline in docking energy, indicating that our method is robust and effective in generating molecules with enhanced docking properties. The random baseline was achieved by sampling molecules from the available chemical space (Enamine building blocks and available reactions) and docking them to the binding sites. These results underline the efficacy of our model in optimizing molecular design for better binding performance.}

\paragraph{Comparison Against Existing Methods}{
Table \ref{tab:comparemodels} presents a comparative analysis of molecule generation using synthesis models (ChemProjector \cite{luo2024projectingmoleculessynthesizablechemical} and SynNet \cite{synnet}), ligand-based 3D models (LigDream \cite{ligdream} and SQUID \cite{SQUID}), and SynthFormer, for the task of generating 100 molecules for each target. ChemProjector is inherently designed for molecule-to-molecule tasks, which naturally results in the generation of analogs. The same applies to SynNet, making direct comparisons less fair. In contrast, 3D methods are guided only by scoring functions, and as demonstrated by SynFlowNet, the SA score is not a discriminatory feature for molecules. Nonetheless, this underscores the importance of SynthFormer as a model specifically tailored for synthesizable ligand-based design.}

Synthesis scores for 3D models were taken from Drug Pose \cite{Drugpose}, where they queried Enamine Real Space for existing molecules. For other models, the scores were based on the number of generated molecules with available synthetic pathways.

Luo et al. demonstrate strong performance in generating molecules similar to known actives, as evidenced by the low $\Delta$ Dock (2.55) and Tanimoto (0.53) scores.$\Delta$Dock (2.55) represents the average difference between the docking score of the reference compound and the docking scores of the generated compounds. These results suggest the generated molecules closely resemble existing drugs and exhibit favorable binding affinities. This observation highlights a limitation of Chemprojector and Synnet in a drug discovery context, where identifying molecules with new scaffolds is crucial for expanding the pool of potential drug candidates. SynthFormer, on the other hand is able to produce similarly active compounds as Chemprojector with low similarity. 

In comparison to 3D methods, SynthFormer achieves a $\Delta$ Dock score of 2.46, which is slightly better than Ligdream (2.78) and comparable to SQUID (2.39), suggesting that SynthFormer generates molecules with docking performance similar to existing methods. In terms of chemical diversity, SynthFormer outperforms both SQUID and Ligdream with a significantly lower $\Delta$ Tanimoto score of 0.09, compared to SQUID (0.24) and Ligdream (0.22). This indicates that the molecules generated by SynthFormer are more distinct from known compounds. Furthermore, SynthFormer achieves a synthesisability of 100\%, far exceeding the synthesisability of SQUID (23.6\%) and Ligdream (32.4\%), demonstrating that all molecules generated by SynthFormer are synthetically feasible. These results highlight the advantages of SynthFormer in generating diverse, chemically distinct, and synthetically accessible molecules.

\subsection{Other Evaluations}

We present additional capabilities of SynthFormer, that allow hit expansion and compound optimization, and we show evidence of meaningful encoding space organization due to effective training. None of these tasks have existing benchmarks for this particular framework and thus we do not compare against other compounds.

\textbf{Building Block Encoding Exploration. }
To further investigate the structural similarity between building blocks, we leveraged the building block encodings produced in our synthesis aware decoder module denoted as Z'. Specifically, we extracted these encodings and used cosine similarity to quantify the closeness of different building blocks.

We revealed that the encodings captured these structural features with remarkable fidelity, observing that similar building blocks not only shared comparable ring configurations but also maintained consistent atom type counts.

Moreover, we selected 100 building blocks and identified the 10 most similar ones based on cosine similarity (see Appendix A, Figure \ref{fig:buildingblock similarity}), finding that top 30\% of the building blocks had an average Tanimoto score of 0.67. This shows that the synthesis-aware decoder effectively captures structural nuances while preserving essential chemical features.

\textbf{Hit Expansion. }
 Our model is also effective for hit expansion in drug discovery, providing a strategic method for identifying structurally similar and synthesizable analogs of hit molecules \citep{Keseru2006, levin2023}. To demonstrate this, we applied our model to design inhibitors for all 10 proteins that have been used for the active ligand design task, following the setup from \citet{levin2023}. We constrained the generation process by initializing with the original structure and then sampling further for possible modifications. Our objective is to identify potentially active synthesizable molecules that are close to the molecule from the crystal structure.

To achieve this, we encode each crystallized molecule as a Morgan fingerprint and input the [Start] token along with the Morgan fingerprint. We then sample an additional building block and the relevant reactions necessary to grow the molecules. A total of 100 molecules are sampled (see Appendix A Figure \ref{fig:SynthFormer_hit_molecules})  and docked back to the relevant protein structures. Afterward, we select the pose with the best score and compare it against the score of the reference ligand.

We generated ~100 analogs for each seed molecule, resulting in a diverse set of molecules with varying scores and structural similarities. Impressively, 11.4\% of these analogs surpassed the original hit in terms of energy, while maintaining an average Tanimoto similarity of 0.67.

\textbf{Molecule Optimization. }
We implement a genetic algorithm approach (See Appendix B) for molecular optimization within a synthetic reaction tree framework, aiming to discover molecules with improved properties. In this part we enhance molecular properties by altering reagents in the reaction steps. For each of the 10 molecules in the Protein Data Bank (PDB), we generate new molecules and using cosine similarity we adjust the reagents in the predicted list to obtain the desired molecules and we resample the reactions. This process helps identify pathways that yield the most optimal molecules based on targeted properties.

Once modifications are applied, we re-dock the resulting molecules, compare their properties, such as binding affinity and physicochemical properties (e.g., logP, druglikeness). By iterating through cycles of modification and evaluation, we quantify improvements in properties over 3 cycles. We generate new molecules, score them, pick the top molecules, and repeat this process 3 times. We show that we can decrease the logp by 0.21 and for a group of molecules druglikeness increase 0.09, while the average energy increases by 0.03 kcal/mol (worse).
\vspace{-0.2cm}
\section{Conclusions}
\vspace{-0.15cm}
SynthFormer presents a novel approach in drug discovery by effectively generating synthetically accessible molecules directly from pharmacophores. Through its integration of a 3D equivariant GNN and a synthesis-aware decoding mechanism, it addresses the critical challenge of designing both active and synthesizable molecules. The results demonstrate that SynthFormer not only outperforms random baselines in docking performance but also achieves higher synthesisability compared to existing 3D models. These findings highlight the model’s potential to accelerate drug discovery by optimizing molecular design and exploring the synthesizable chemical space around hit compounds, marking it as a promising tool for advancing the field. Moreover, it has shown effectiveness in hit expansion and molecule optimization, further expanding its applicability in drug discovery workflows.
\bibliography{main}
\bibliographystyle{apalike}
\newpage

\section{Appendix}

\begin{figure}[H]
    \centering
    \begin{subfigure}
        \centering
\includegraphics[width=\linewidth]{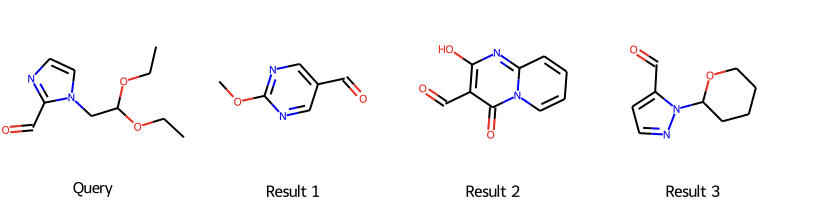}
        \label{fig:screen1}
    \end{subfigure}
    \begin{subfigure}
        \centering
        \includegraphics[width=\linewidth]{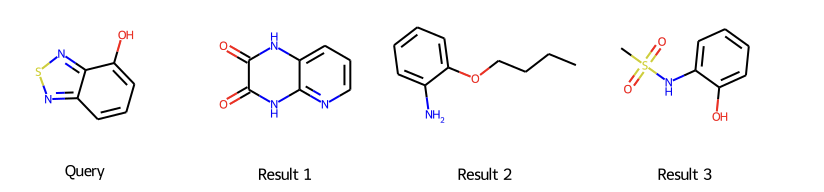}
        \label{fig:screen2}
    \end{subfigure}
    \begin{subfigure}
        \centering
        \includegraphics[width=\linewidth]{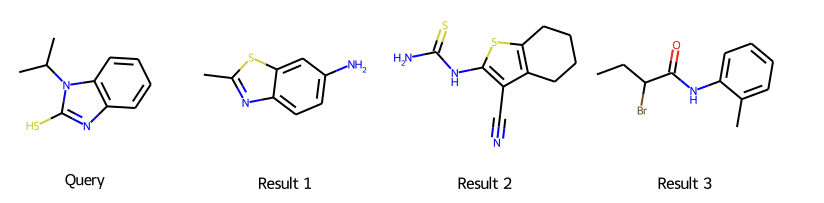}
        \label{fig:screen3}
    \end{subfigure}
    \caption{The building block encoding of the query molecule serves as the reference, with the three closest molecules identified using cosine similarity, preserving significant structural similarity.}
    \label{fig:buildingblock similarity}
\end{figure}

\begin{figure*}
    \centering
    \includegraphics[width=\textwidth]{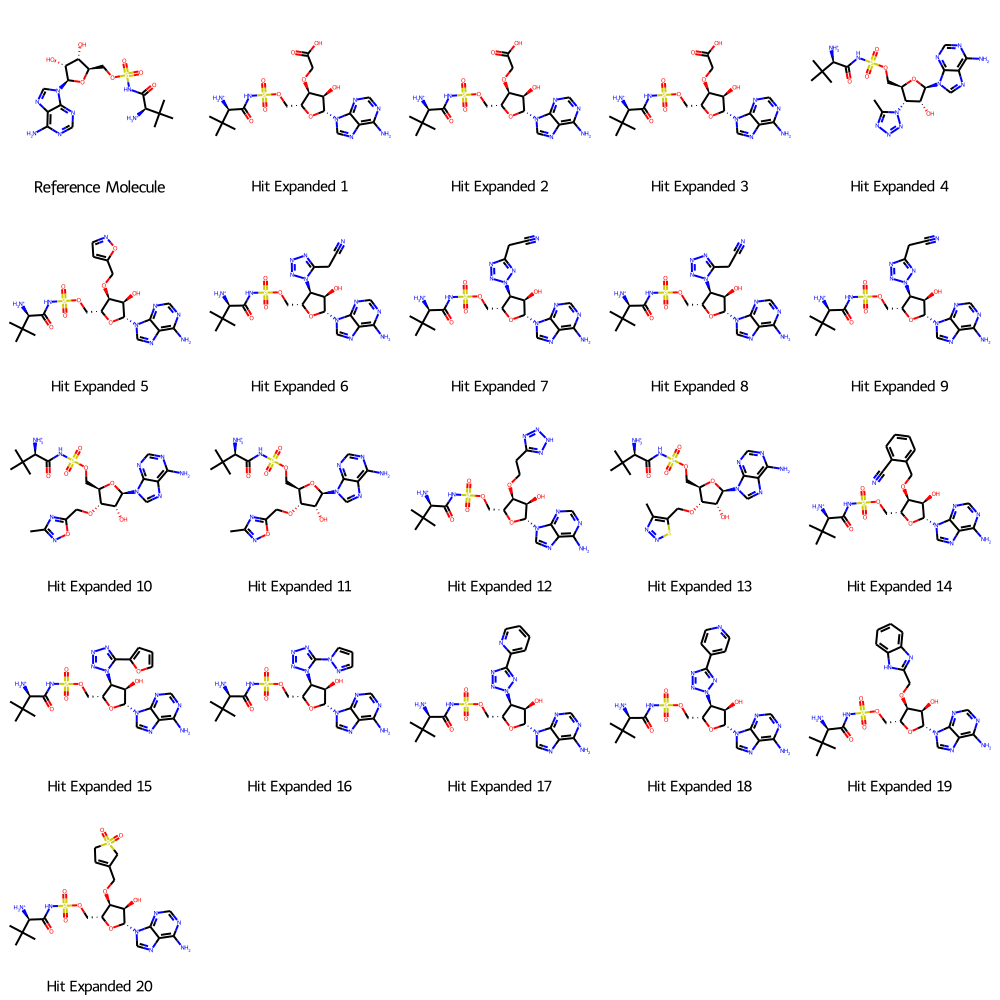}  
    \caption{The first molecule corresponds to the reference structure derived from the 3COY PDB ID. The subsequent molecules are expanded hits generated by the SynthFormer model, demonstrating structural diversity and novel chemical scaffolds.}
    \label{fig:SynthFormer_hit_molecules}
\end{figure*}



\end{document}